\documentclass[10pt,twocolumn,letterpaper]{article}

\usepackage{wacv}
\usepackage{times}
\usepackage{epsfig}
\usepackage{graphicx}
\usepackage{amsmath}
\usepackage{amssymb}

\DeclareMathAlphabet{\pazocal}{OMS}{zplm}{m}{n}
\newcommand{\Lb}{\pazocal{L}}
\usepackage{enumitem}
\usepackage{cite}   
\usepackage{enumitem}
\usepackage{graphicx}
\usepackage{subfloat}
\usepackage{tabularx}
\usepackage{adjustbox}
\usepackage{amssymb}
\usepackage{multirow}
\usepackage[flushleft]{threeparttable}
\usepackage{subfigure}

\newcommand{\figref}[1]{Fig.~\ref{#1}}
\newcommand{\tabref}[1]{Table.~\ref{#1}}
\newcommand{\eqnref}[1]{Eq.~(\ref{#1})}
\newcommand{\secref}[1]{Sec.~\ref{#1}}

\wacvfinalcopy 

\def\wacvPaperID{317} 

\ifwacvfinal\fi
\begin{document}

\title{Learning Image Representations by Completing Damaged Jigsaw Puzzles}

\author{\hspace{-0.1in}Dahun Kim\\
\hspace{-0.1in}KAIST\\
\hspace{-0.1in}{\tt\small mcahny@kaist.ac.kr}
\and
\hspace{-0.1in}Donghyeon Cho\\
\hspace{-0.1in}KAIST\\
\hspace{-0.1in}{\tt\small cdh12242@gmail.com}
\and 
\hspace{-0.1in}Donggeun Yoo\\
\hspace{-0.1in}KAIST\\
\hspace{-0.1in}{\tt\small dgyoo@rcv.kaist.ac.kr}
\and
In So Kweon\\
KAIST\\
{\tt\small iskweon@kaist.ac.kr}
}

\maketitle
\ifwacvfinal\thispagestyle{empty}\fi

\begin{abstract}
In this paper, we explore methods of complicating self-supervised tasks for representation learning. That is, we do severe damage to data and encourage a network to recover them. First, we complicate each of three powerful self-supervised task candidates: jigsaw puzzle, inpainting, and colorization. In addition, we introduce a novel complicated self-supervised task called ``Completing damaged jigsaw puzzles" which is puzzles with one piece missing and the other pieces without color. We train a convolutional neural network not only to solve the puzzles, but also generate the missing content and colorize the puzzles. The recovery of the aforementioned damage pushes the network to obtain robust and general-purpose representations. We demonstrate that complicating the self-supervised tasks improves their original versions and that our final task learns more robust and transferable representations compared to the previous methods, as well as the simple combination of our candidate tasks. Our approach achieves state-of-the-art performance in transfer learning on PASCAL classification and semantic segmentation.
\end{abstract}

\section{Introduction}

The goal of representation learning is to learn robust and general-purpose visual features. Typically, the amount of labeled data decreases as the extent of annotation increases. The networks trained on limited amount of labeled data are easily overfitted and have poor representation ability. Representation learning is used to avoid this problem by pretraining visual features on large-scale data before training on target tasks.

\begin{figure}[t]
\def\arraystretch{0.5}
\begin{tabular}{@{}c@{\hskip 0.01\linewidth}c@{\hskip 0.01\linewidth}c@{\hskip 0.01\linewidth}}

\includegraphics[width=1.0\linewidth]{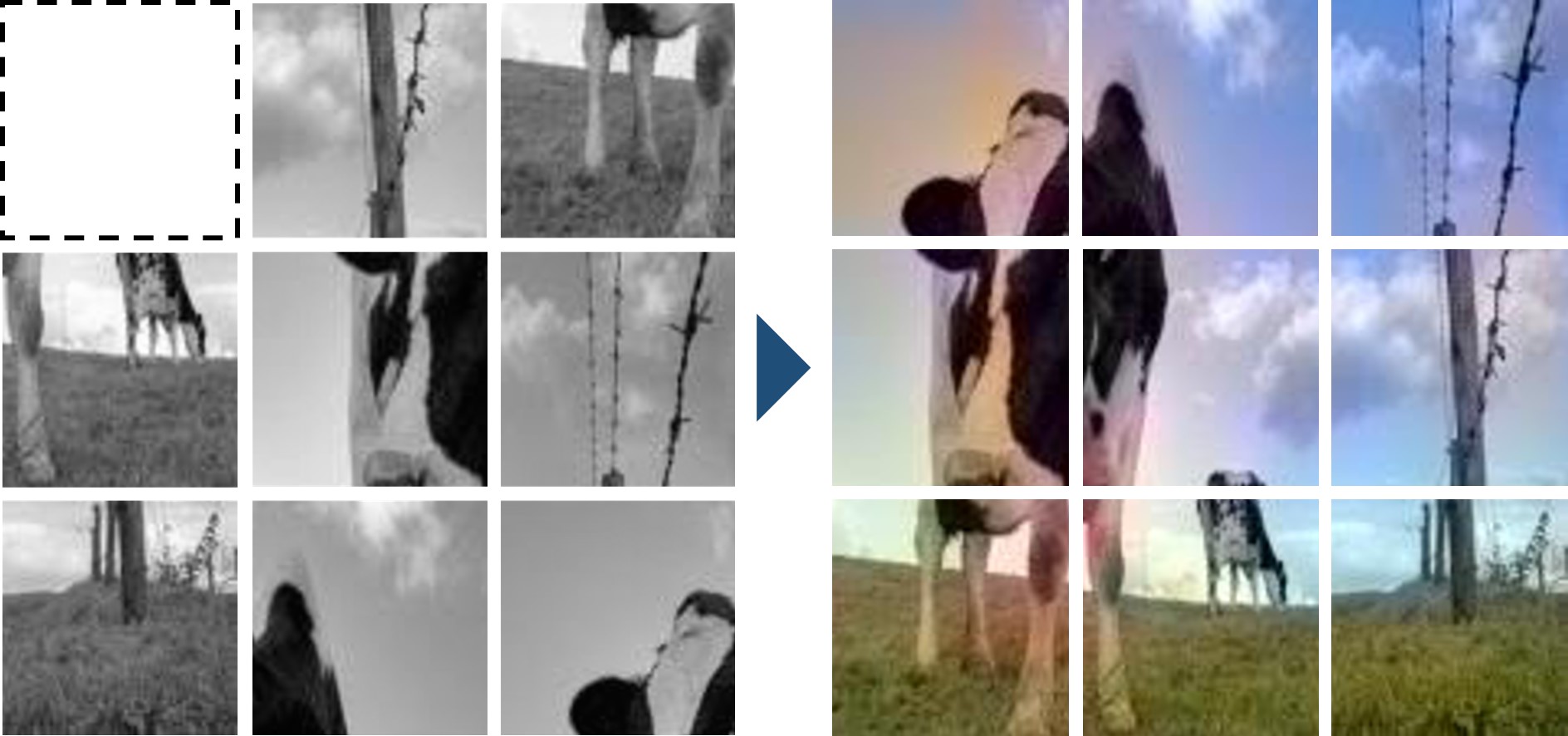}&\\
{\small (a) \hspace{4.2cm}\small (b)}\\
\end{tabular}

\caption{ {\bf Learning image representations by completing damaged jigsaw puzzles. } We sample 3-by-3 patches from an image and create damaged jigsaw puzzles. (a) is the puzzles after shuffling the patches, removing one patch, and decolorizing. We push a network to recover the original arrangement, the missing patch, and the color of the puzzles. (b) shows the outputs; while the pixel-level predictions are in \textit{ab} channels, we visualize with their original $L$ channels for the benefit of the reader.}
\label{fig:teaser}
\end{figure}

Conventional yet still popular method to learn such features is to pre-train image classification~\cite{krizhevsky2012imagenet,Simonyan14c,Googlenet,resnet} on millions of human-labeled data such as ImageNet~\cite{ILSVRC15}. It provides powerful representations and image priors when the target task and data are similar. However, the dependency on human supervision of this traditional method limits its scalability and adaptability to dissimilar target tasks and domains(\textit{e.g.} depth prediction).

Many researches have been conducted to minimize human supervision in computer vision. For example, weakly-supervised learning~\cite{Oquab15CVPR,WSDDN16CVPR,Context16ECCV,self17CVPR,TPL17ICCV} has been proposed to learn object localization using \textit{weak} image-level annotations rather than bounding boxes or pixel-level annotations. In the same vein, recent representation learning has also been improved to minimize human supervision. The emerging family of such methods is self-supervised learning; It manufactures a supervised task and labels from raw images, so that unlimited amount of labeled data can be used. A considerable number of such methods ~\cite{Doersch15ICCV,Noroozi16ECCV,Pathak16CVPR,Zhang16ECCV,Larsson17CVPR,Zhang17CVPR,Wang15ICCV,Lee17ICCV,Donahue17ICLR,Noroozi17ICCV,Wang17ICCV,Doersch17ICCV} have been proposed in last few years. They often train a network to infer geometrical configuration~\cite{Doersch15ICCV,Noroozi16ECCV}, recover missing pixels~\cite{Pathak16CVPR} or channels~\cite{Zhang16ECCV,Zhang17CVPR,Larsson17CVPR} of images. The features learned by these methods have been successfully transferred to different target tasks, such as classification, detection, and semantic segmentation, and resulted in promising performances.

The common intuition of these approaches is that a network obtains useful representations of scenes and objects while struggling to solve a challenge task that requires high-level reasoning. Based on this idea, we propose a concept of complicating a self-supervised task where we raise the difficulty of the task. More specifically, we design more difficult versions of jigsaw puzzle, inpainting, and colorization tasks. We investigate the effectiveness of our approach by transferring the learned features on PASCAL VOC classification, detection, and segmentation tasks~\cite{pascal-voc-2007,pascal-voc-2012}. In order to further the idea, we design a task called ``Completing damaged jigsaw puzzle", which is puzzles with one piece missing and the other pieces without color. Then, jigsaw puzzle, inpainting, and colorization tasks are jointly optimized. The network learned in this way preserves better feature representations for classification, detection and semantic segmentation. 

In summary, our main contributions are as follows:
\begin{itemize}
\item We propose an approach of making self-supervised tasks more challenging for representation learning.
\item We design a problem of completing damaged jigsaw puzzles where three different
self-supervised tasks are complicated and incorporated simultaneously.
\item We show that the representations learned by our approach achieve state-of-the-art performances on PASCAL classification and semantic segmentation~\cite{pascal-voc-2007,pascal-voc-2012} when transferred on AlexNet, compared to existing self-supervised learning methods.
\end{itemize}

\section{Related works}
A considerable number of unsupervised learning approaches have been studied to learn image representations without relying on human-annotation. The most fundamental example is the autoencoder~\cite{Vincent10JMLR}, which is a generative model that reconstructs the input data, aiming to extract the data representation. Since then, various generative models rooted in the autoencoder have been proposed. For example, DCGAN~\cite{Radford16ICLR} and variational auto-encoders~\cite{Kingma14ICLR} have been proposed for further photorealistic reconstruction and feature learning. 

Our study falls into \textit{self-supervised learning} which has emerged as a new stream of unsupervised learning. This technique manufactures supervision signal from the raw visual data and achieves promising results in learning discriminative features. Recent methods commonly use images~\cite{Zhang16ECCV,Zhang17CVPR,Doersch15ICCV,Pathak16CVPR,Larsson17CVPR,Noroozi16ECCV,Noroozi17ICCV,Donahue17ICLR}, and often video~\cite{Wang15ICCV,Pathak17CVPR,MisraZH16ECCV,Isola16ICLRW}, or other sensory data such as egomotion and sound~\cite{AgrawalCM15ICCV, Jayaraman17IJCV,Andrew16ECCV,Relja17ICCV}.

Different supervision signals encourage the network to pay attention to different characteristics in images. Thus, the virtues of the learned representations also differ across the self-supervised tasks. Recent methods on self-supervised feature learning can be broadly categorized according to the type of knowledge preferred in the training: spatial configuration, context, and cross-channel relations.

\bigskip\noindent\textbf{Spatial Configuration. } The methods that operate on the spatial dimension of images usually extract the patches from the image and learn the network to infer spatial relations between them. Doersch~\etal~\cite{Doersch15ICCV} proposed a problem with 3-by-3 puzzles, where the network sees one of the outer patches, and predicts its relative position to the center patch. Noroozi and Favaro~\cite{Noroozi16ECCV} learn image representations by solving the jigsaw puzzle with the 3-by-3 patches which imposes a challenging task of estimating what permutation has been used in shuffling. The learned features well capture the geometrical configuration of the objects as mentioned in~\cite{Doersch15ICCV}.

\bigskip\noindent\textbf{Image Context. } A contextual autoencoder was proposed by Pathak~\etal~\cite{Pathak16CVPR} in order to drive representation learning. The supervisory signal comes from inpainting task where the network is encouraged to recover dropped part of the image from the surrounding pixels. Also, Isola~\etal~\cite{Isola16ICLRW} exploited a co-occurance cues as a self-supervision where the network takes two isolated patches and predict whether or not they were taken from nearby locations in an image. These methods allow the network to learn contextual relations between part of an image and the rest or between each object parts/instances in an image.

\bigskip\noindent\textbf{Cross-Channel Relations. } The methods that manipulate the images in channel domain have also been proposed. Typically, they remove one subset of the image channels, and train the network to recover it from the remaining channel(s). Zhang~\etal~\cite{Zhang16ECCV} and Larsson~\etal~\cite{Larsson17CVPR} obtain self-supervision from the task of colorization where the network predicts \textit{ab} channels given L channel. Zhang~\etal\cite{Zhang17CVPR} took a one step further by learning colorization together with the inverse mapping from \textit{ab} channels to L channel.
\begin{figure*}[]
\centering
\def\arraystretch{1.0}
\begin{tabular}{@{}c@{\hskip 0.05\textwidth}c@{\hskip 0.05\textwidth}c}
\includegraphics[width=0.3\textwidth]{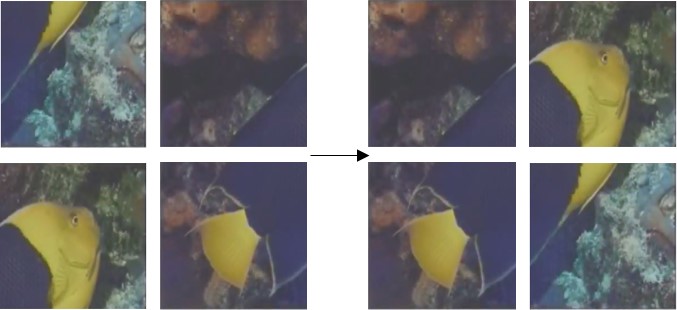} &
\includegraphics[width=0.3\textwidth]{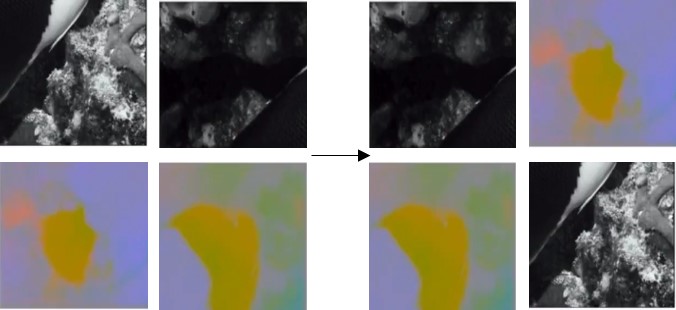} &
\includegraphics[width=0.3\textwidth]{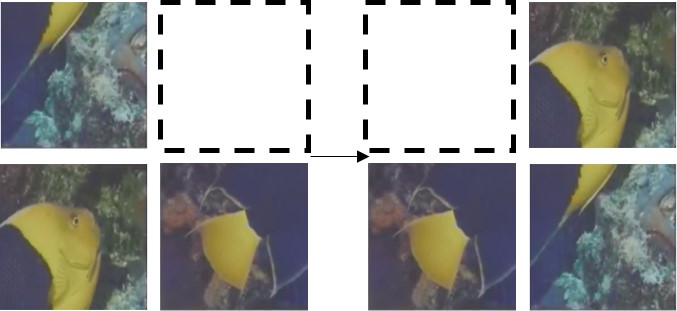}\\
{\small (a)} & {\small (b)} & {\small (c) }\\
\end{tabular}

\begin{tabular}{@{}c@{\hskip 0.05\textwidth}c@{\hskip 0.05\textwidth}c@{\hskip 0.05\textwidth}c}
\includegraphics[width=0.21\textwidth]{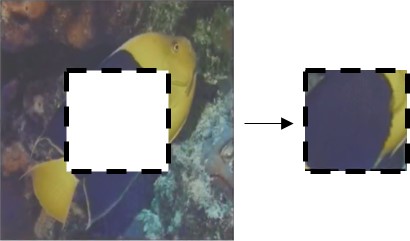} &
\includegraphics[width=0.21\textwidth]{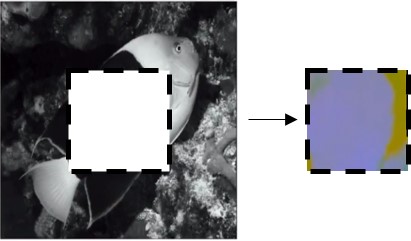} &
\includegraphics[width=0.25\textwidth]{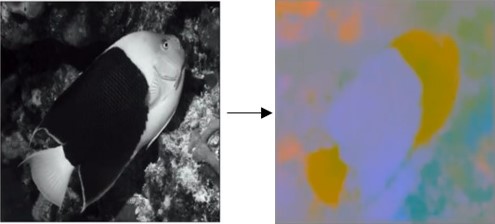} &
\includegraphics[width=0.18\textwidth]{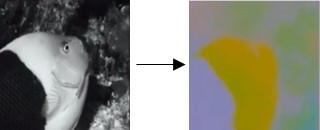}\\
{\small (d)} & {\small (e)} & {\small (f) } & {\small (g)}\\
\end{tabular}

\caption{{\bf Illustrations of complicating self-supervised tasks. }(a) Conventional 2$\times$2 jigsaw puzzles. (b) Complicated 2$\times$2 jigsaw puzzles; each patch's $L$ or $\textit{ab}$ channel is dropped. (c) Complicated 2$\times$2 jigsaw puzzles; one of the patches is completely dropped. (d) Conventional inpainting. (e) Complicated inpainting; it outputs in $\textit{ab}$ channels from an input in only $L$ channel. (f) Conventional colorization. (g) Complicated colorization; only one-quarter of the entire image is given for colorization.}
\label{fig:Each Tasks}
\end{figure*}


\bigskip\noindent\textbf{Combining Multiple Self-supervised Tasks.} The aforementioned methods are essentially relying on a single supervisory signal. Recently, representation learning by multiple supervisory signals has also emerged. Zhang~\etal~\cite{Zhang17CVPR} proposed a bidirectional cross-channel prediction to aggregate complementary image representations. They propose a network split into to two groups, and each subnetworks are trained separately. Wang~\etal~\cite{Wang17ICCV} exploited two self-supervised approaches to unify different types of invariance appearing in the two approaches. Doersch and Zisserman~\cite{Doersch17ICCV} combine multiple self-supervised tasks to create a single universal representation. However, each of the methods have limitations. In ~\cite{Zhang17CVPR}, splitting the network reduces the number of parameters by half which might limit the feature transferability. Also, ~\cite{Wang17ICCV} trains two tasks in sequential order. That is, the training on ranking video frames~\cite{Wang15ICCV} comes only after the training on estimating relative position~\cite{Doersch15ICCV} finishes. Lastly, the involved tasks in ~\cite{Doersch17ICCV} operate on very different inputs, which hinders simultaneous training of all tasks and requires special handlings. 

Our study shares the goal with ~\cite{Zhang17CVPR, Wang17ICCV} and ~\cite{Doersch17ICCV} where we want to learn representations that have all-round capability in every downstream task. However, our approach differs in the strategy; We squeeze a network to solve more complicated tasks, and in the same vain, our final method combines the complicated tasks and trains them simultaneously.

\section{Approach}

A number of recent self-supervised learning methods commonly operate via \textit{damage-and-recover} mechanisms. In other words, the networks are supervised to recover intentionally damaged image data. For the purpose of representation learning, the damages are designed so that the recovery requires the high-level understanding of the objects and the scene. During training, the representations that are necessary for the recovery are learned, resulting in task/damage-specific features. For example, the spatial configuration is damaged in jigsaw puzzles, so the learned representations are focused on the configuration and geometry of objects. Similarly, the representations learned from inpainting and colorization preferably encode contextual and cross-channel relations as analyzed in~\cite{Pathak16CVPR} and ~\cite{Larsson17CVPR}, respectively. 

Motivated by the mechanism above, we design a strategy where we drive the network to recover even more severe damage. More specifically, we do further damage to the data in jigsaw puzzle, inpainting, and colorization to make them more challenging as illustrated in~\figref{fig:Each Tasks}. The methods of complicating each of the tasks are explained in \secref{sec:Each Tasks}. Furthermore, in order to maximize the effectiveness of our approach, we incorporate those three tasks in a single problem, ``Completing damaged jigsaw puzzles", as detailed in~\secref{sec:Damaged Puzzles}.

\begin{figure*}[t]
\centering
\includegraphics[width=0.98\textwidth]{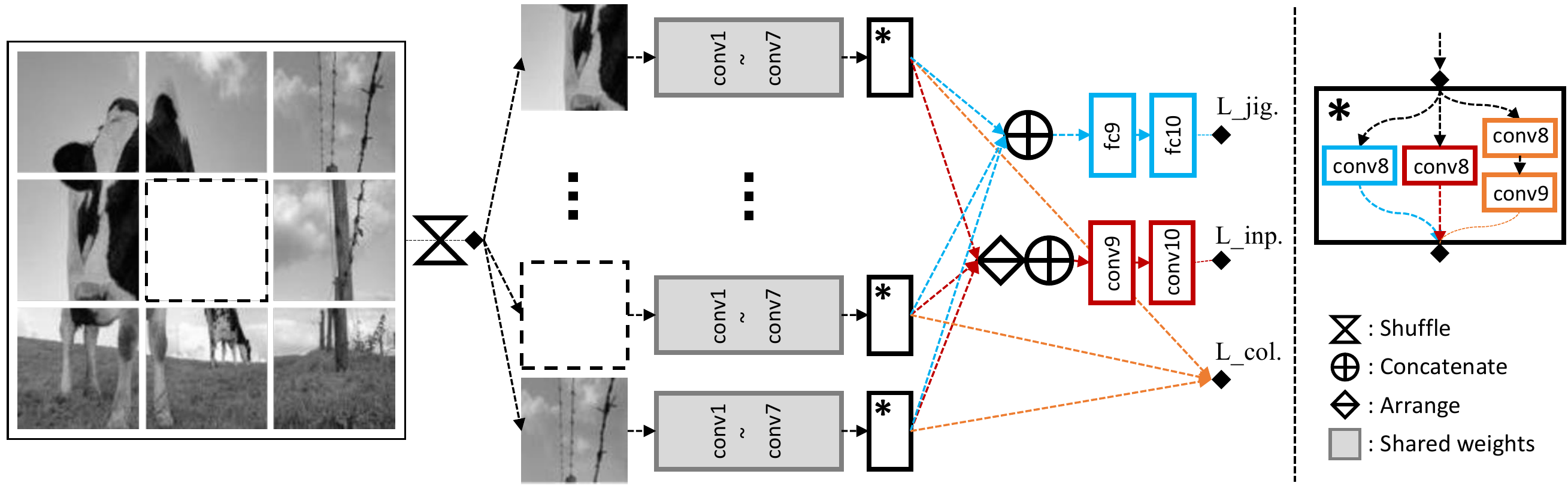}
\caption{{\bf The architecture for ``Completing damaged jigsaw puzzles". } It is a 9-tower siamese network. The shared tower(colored in gray) consists of AlexNet $conv1$-$7$ layers. note $fc6$-$7$ are converted into equivalent $conv6$-$7$ layers for the pixel-level outputs. The task branches for jigsaw puzzle, inpainting, and colorization are marked in blue, red, and orange, respectively. The learned shared tower is used for transfer learning on downstream tasks.}
\label{fig:architecture}
\end{figure*}

\subsection{Complicating Each Self-supervised tasks}
\label{sec:Each Tasks}
In this section, we briefly review each of jigsaw puzzle, inpainting and colorization, and explain the methods of complicating them. Considering that different damages teach different lessons, we do additional damage to the data domains that have remained intact in the original task. The effectiveness of the complicated versions is quantitatively evaluated in \secref{sec: fine-tuning pascal}.

\bigskip\noindent\textbf{Jigsaw Puzzle. } With 2-by-2 puzzles, let us define $S$ a sequence of puzzle patches $X\textsubscript{1}$-$X\textsubscript{4}$ shuffled by a permutation $P$. The spatial configuration of objects is intermixed by the permutation.

Accordingly, we consider two additional types of damage that make jigsaw puzzles more difficult. First, we do damage in the channel-wise domain, where half of the puzzles have only the $L$ channel and the other half, \textit{ab} channels, as shown in \figref{fig:Each Tasks}-(b). Successfully solving the puzzles requires not only the knowledge on spatial configuration, but also the understanding of the cross-channel relations between $L$ and \textit{ab} channels. Second, we damage the image context by removing one piece from a complete set of puzzles, as shown in~\figref{fig:Each Tasks}-(c). In practice, a piece is discarded with a probability of 0.4 and the missing contents are replaced with Gaussian noise. Doing well on this task may require extra understanding on the full context without seeing the missing area. 

As in~\cite{Noroozi16ECCV}, we train an AlexNet-based network to learn a mapping $\hat{P} = f\textsubscript{jig}(S)$ to a probability distribution over 24(that is, 4!) possible permutations $\hat{P}\in[0,1]\textsuperscript{24}$ with loss, 

\begin{eqnarray}
\Lb_{jig} =  -\sum & P\log(\hat{P}) .
\label{equ:jigsaw_loss}
\end{eqnarray}

\bigskip\noindent\textbf{Inpainting. } 
Inpainting is a problem of restoring lost regions of an image. In the field of representation learning, a small patch $X_p$  is removed from the image $X$, and remaining parts $X_r$ are used for inferring the removed patch $X_p$. It is formulated as 
\begin{eqnarray}
\hat{X_p} = f\textsubscript{inp}(X_r).
\label{equ:inpainting_eq}
\end{eqnarray}
By solving this problem, the network learns contextual information of $X_r$ and between $X_r$ and $X_p$.


In order to do damage of a different flavor, we discard a subset of image channels. Unlike the original inpainting where all channels are given (\figref{fig:Each Tasks}-(d)), our complicated inpainting requires generation of \textit{ab} channels of the missing region from the surrounding pixels in $L$ channel (\figref{fig:Each Tasks}-(e)). While struggling to solve this problem, the network learns cross-channel relations as well as the contextual information. We use Euclidean distance between the prediction and the ground truth as a loss as proposed in~\cite{Pathak16CVPR} as

\begin{eqnarray}
\Lb_{inp} =  \left \|  \hat{X_p^{ab}} - f\textsubscript{inp}(X_r^{L})  \right\|_{2}^{2},
\label{equ:inpainting_loss}
\end{eqnarray}
where superscripts $L$ and \textit{ab} denote the input's $L$ and \textit{ab} channels, respectively.

\bigskip\noindent\textbf{Colorization. } Colorization and other cross-channel prediction tasks~\cite{Zhang16ECCV, Zhang17CVPR, Larsson17CVPR} discard and recover a subset of image channels to learn cross-channel relations.

Additional damage for more difficult colorization takes place in the context domain. We encourage the network to see only part of images, and colorize in the absence of the full context. Specifically, we feed the network with the $L$ channel of only one patch out of the 2$\times$2 puzzles, and push it to colorize the patch as shown in~\figref{sec:Each Tasks}-(g). The colorization becomes more difficult since only one-quarter of the entire image is available. 

As in~\cite{Zhang16ECCV}, the network learns a mapping $\hat{{X^{\textit{ab}}}}$ = $f_{col}(X^{L})$ to a probability distribution over possible colors $\hat{{X^{\textit{ab}}}} \in [0,1]^{313}$, where the $\textit{ab}$ are quantized to 313 values. We train the network with the 313-way classification loss as, 

\begin{eqnarray}
\Lb_{col} =  -\sum v({X^{\textit{ab}}} \log(\hat{{X^{\textit{ab}}}})) ,
\label{equ:color_loss}
\end{eqnarray}
where v($\cdot$) denotes a color-class rebalancing term.


\subsection{Completing Damaged Jigsaw Puzzles}
\label{sec:Damaged Puzzles}

In order to further develop our idea, we design our final problem, ``Completing damaged jigsaw puzzles", by involving all the damages and recoveries mentioned above.
As its name indicates, this problem requires the simultaneous recovery of the following damages: (1) shuffling the image patches, (2) discarding one patch, (3) dropping $\textit{ab}$ channels in all the patches. During training, the network is encouraged to arrange the puzzles, recover the missing context, and colorize the patches. In practice, recovering the missing patch is defined as generating $\textit{ab}$ channels of the missing region from the surrounding pixels in $L$ channel.

Recent self-supervised learning methods that use multiple self-supervised tasks either assign separate features to each tasks~\cite{Zhang17CVPR}, train each tasks in sequential order~\cite{Wang17ICCV}, or jointly train the tasks~\cite{Doersch17ICCV}. We share with them the goal of learning a single set of well-rounded representations. However, our approach complicates each involved tasks to fuel the \textit{damage-and-recover}, whereas the previous methods adopt the original form of existing self-supervised tasks. More specifically, our final problem involves a jigsaw puzzle with one piece missing, inpainting across channels, and colorization with a narrower view, which are more complicated than their predecessors. Also, each tasks are intermingled in a way that some tasks share the knowledge. That is, the understanding of cross-channel relation supports both the colorization and the inpainting, and the contextual information is shared across all tasks. As a result, the network learns to effectively integrate and propagate the different knowledge on the spatial configuration, image context, and cross-channel relations into the final representations. Finally, all our involved tasks share the input space: a set of damaged puzzles. This makes our approach immune to the risk in ~\cite{Zhang17CVPR,Doersch17ICCV} that use different inputs for each tasks, where the network might task-specifically encode the representations depending on the type of inputs, as stated in~\cite{Doersch17ICCV}.

In practice, our method operate on 3$\times$3 puzzles rather than 2$\times$2, for more discriminative representations.


\bigskip\noindent\textbf{Architecture and Losses.} Our architecture is shown in \figref{fig:architecture}. It is a 9-tower siamese network as in \cite{Noroozi16ECCV}. The shared tower follows the standard AlexNet\cite{krizhevsky2012imagenet} to provide a fair comparison with recent self-supervised learning methods \cite{Zhang16ECCV, Zhang17CVPR,Noroozi16ECCV,Noroozi17ICCV,Doersch15ICCV,Pathak16CVPR,Donahue17ICLR,Larsson17CVPR,Wang15ICCV,Wang17ICCV}. The task branches of the jigsaw puzzle, inpainting, and colorization are rooted to the shared tower, and colored in blue, red, and orange, respectively. 

In the jigsaw branch, 9 sets of the common features ($conv7$ features) pass through a fully-connected layer, $fc8$(blue), and are concatenated, then fed into two more fully-connected layers up to $fc10$(blue), resulting in a 1000-long vector. We use the same $\Lb_{jig}$ as \eqnref{equ:jigsaw_loss}. In the inpainting branch, the 9 features go through a 1$\times$1 convolutional layer. This time, we arrange the features before concatenating them as we know what permutation has been used in the inputs. After two more 1$\times$1 convolutions($conv9$, $conv10$, red), the features have a volume of $7\times7\times313$, where 313 denotes the number of quantized color values as in \cite{Zhang16ECCV}. Note that we use a classification loss rather than \eqnref{equ:inpainting_loss} as,
\begin{eqnarray}
\Lb_{inp}^{cls} =  -\sum v({X_p^{\textit{ab}}} \log(\hat{{X_p^{\textit{ab}}}})) ,
\label{equ:inpainting_cls_loss}
\end{eqnarray}
where $\hat{X_p^{\textit{ab}}}$ denotes the predicted chromaticity values of the missing puzzle. Each of the 9 features is fed into the colorization branch, resulting in 9 branches. Each branch is an equivalent form of the network in \cite{Zhang16ECCV} which has two more $1\times1$ convolutions($conv8$, $conv9$, orange) after the shared tower, resulting in features of $7\times7\times313$. Our colorization loss is a sum of the 9 losses of \eqnref{equ:color_loss} as,
\begin{eqnarray}
\Lb_{col} =  -\sum_{i=1}^{9}(\sum v({X_{i}^{\textit{ab}}} \log(\hat{{X_{i}^{\textit{ab}}}}))) ,
\label{equ:color_9_loss}
\end{eqnarray}
where $X_{i}$ denotes i\textit{th} of the input patches.
Finally, our loss for ``Completing damaged jigsaw puzzles" is the sum of the three losses as, 
\begin{eqnarray}
\Lb_{final} = \Lb_{jig} + \alpha\Lb_{inp}^{cls} + \beta\Lb_{col},
\label{equ:color_loss}
\end{eqnarray}
where $\alpha$ and $\beta$ are weighting parameters.


\bigskip\noindent\textbf{Simple Combination.} We also consider combining the original forms of self-supervised tasks, conceptually following \cite{Doersch17ICCV}. We jointly train original versions of the three tasks: jigsaw puzzles, inpainting, and colorization. Although the types of involved tasks are different to \cite{Doersch17ICCV}, we provide a self-comparison on the effectiveness of our approach and the simple combination in \secref{sec:comparing}.

\section{Training}
\label{sec: training}

We train our proposed network on 1.3M images from the training set of ImageNet without annotations. We resize the input images to 312$\times$312 pixels, and extracted patches of 140$\times$140 and 85$\times$85, in 2-by-2 and 3-by-3 puzzles, respectively. We use caffe~\cite{jia2014caffe} for implementation. The network is trained by ADAM optimizer~\cite{Kingma15ICLR} for 350K iterations with batch size of 64 on a machine with a GTX 1080-Ti GPU and an intel i7 3.4GHz CPU. The learning rate is set to $10^{-3}$, and is dropped by a factor of 0.1 every 100K iterations. We use $\alpha$, $\beta$ = 0.01 for the experiment in \secref{sec:Damaged Puzzles}. Inpainting and colorization of \secref{sec:Each Tasks} follow the protocol of their original papers~\cite{Pathak16CVPR,Zhang16ECCV}, respectively.


\section{Results and Discussions}

In this section, we provide both quantitative and qualitative evaluations and discussions of our self-supervised learning approach. Further transfer learning results on new tasks(\textit{e.g.} depth prediction) and with deeper network(\textit{e.g.} vgg~\cite{Simonyan14c}) are presented in our supplementary material.

\subsection{Fine-tuning on PASCAL}
\label{sec: fine-tuning pascal}

In this section we evaluate the effectiveness of both the ``Complicating each self-supervised tasks" in \secref{sec:Each Tasks} and our final task, ``Completing damaged jigsaw puzzles" in \secref{sec:Damaged Puzzles}. To do this, we transfer the learned representations to a standard AlexNet~\cite{krizhevsky2012imagenet} and rescale the weights via ~\cite{krahenbuhl16ICLR}. We test on some or all of the PASCAL tasks, using VOC 2007~\cite{pascal-voc-2007} for classification and detection, VOC 2012~\cite{pascal-voc-2012} for segmentation; these are standard benchmarks for representation learning.

\subsubsection{Complicating each self-supervised task}
\label{sec: each task}

In \secref{sec:Each Tasks}, we explore the idea of complicating the jigsaw puzzle, inpainting, and colorization to benefit representation learning. We evaluate the effectiveness of each complications by comparing the performances before and after the complications in downstream tasks: classification and semantic segmentation. 

The results are shown in ~\tabref{table:table1}. In all cases, the complicated self-supervised tasks consistently achieve higher scores than their predecessors both in classification and segmentation. These results indicate that the capacity of the network was still above the difficulty of the existing self-supervised tasks, and that indeed, useful representations can be extracted more via solving more difficult tasks. 

\begin{table}
\resizebox{\columnwidth}{!}{%
\begin{tabular}{l cl cc cc cc}
\hline
\bf Method                   && \bf Complication  &&\bf Class.  &&\bf Segm.  \\
\hline
\hline
Jigsaw(~\secref{sec:Each Tasks})            &&None             &&64.7     &&34.9 \\
Jigsaw(~\secref{sec:Each Tasks})            &&L-or-ab dropped  &&65.5     &&35.7 \\
Jigsaw(~\secref{sec:Each Tasks})            &&A piece removed  &&65.3     &&35.7 \\

\hline
Inpainting~\cite{Pathak16CVPR}                    &&None             &&56.5     &&29.7 \\
Inpainting(~\secref{sec:Each Tasks})              &&Cross-Channel    &&57.7     &&30.2\\

\hline
Colorization~\cite{Zhang16ECCV}                   &&None             &&65.9     &&35.7 \\
Colorization(~\secref{sec:Each Tasks})            &&Narrow view      &&66.7     &&36.8 \\
\hline
\end{tabular}
}
\smallskip
\caption{{\bf Effectiveness of complicating self-supervised tasks on PASCAL.} Classification is evaluated on PASCAL VOC 2007 with testing frameworks from~\cite{krahenbuhl16ICLR}, using mean average precision(mAP) as a performance measure. Segmentation is evaluated on PASCAL VOC 2012 with testing framework from ~\cite{Long15CVPR}, which reports mean intersection over union(mIU). }
\label{table:table1}
\end{table}

\subsubsection{Completing Damaged Jigsaw Puzzles}

\begin{figure}[t]
\def\arraystretch{0.5}
\begin{tabular}{@{}c@{\hskip 0.01\linewidth}c@{\hskip 0.01\linewidth}c@{}}

\includegraphics[width=1.0\linewidth]{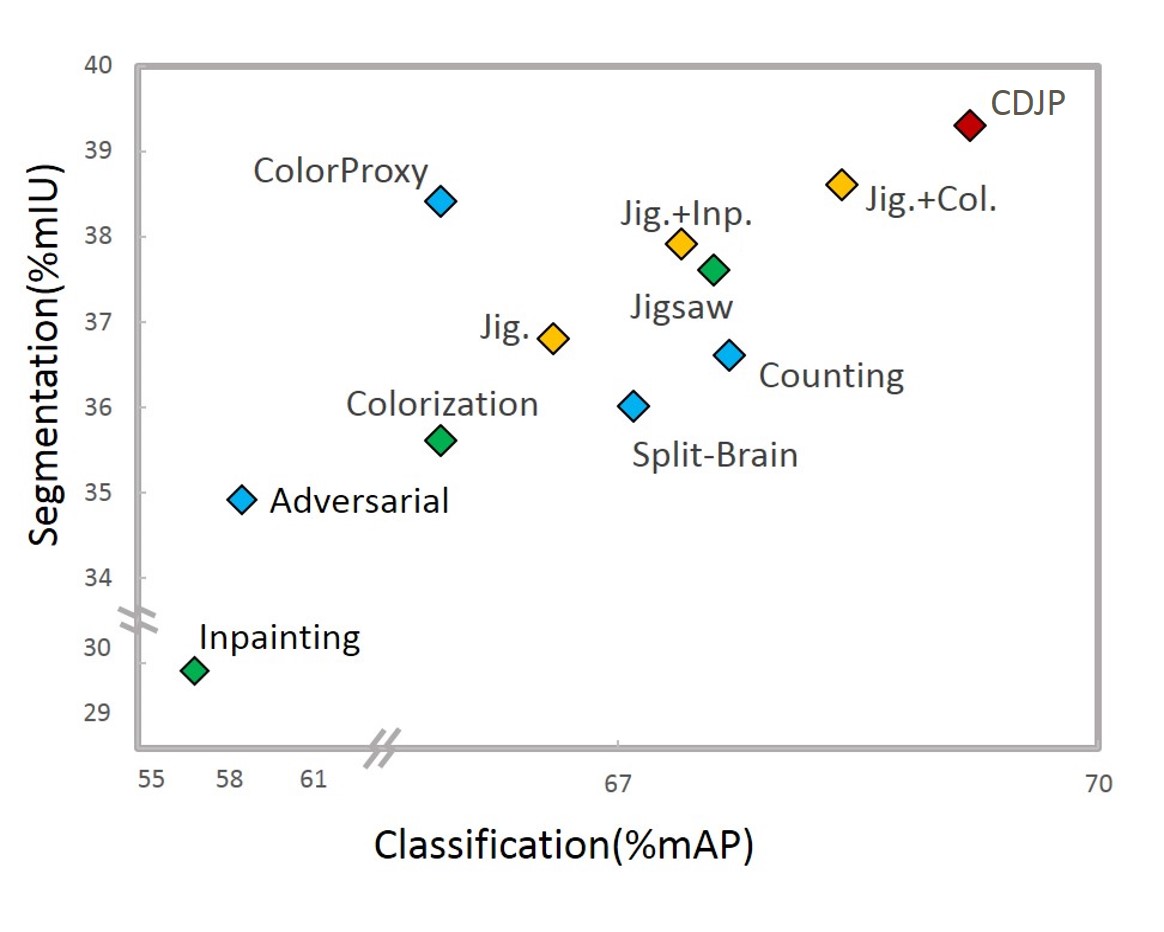}&\\
\end{tabular}

\caption{ {\bf Summarization of performances of different self-supervised learning methods and combinations. } We compare the state-of-the-art methods(\tabref{table:table2}), our final method (CDJP), and each involved tasks in our final method and the simple combination(~\tabref{table:ablation}). The involved tasks, their original versions, the simple combination, the other existing methods, and our final method are marked in orange, green, gray, blue and red, respectively. Note that $Jig.$ is what we reproduced in our architecture.}
\label{fig:cls_vs_seg}
\end{figure}

We evaluate how beneficial is our final self-supervised task, ``Completing damaged jigsaw puzzles", in learning representations. We transfer the learned weights from the shared tower~\figref{fig:architecture} on classification, detection, and semantic segmentation. As shown in ~\tabref{table:table2}, our method outperforms all the previous methods in classification and segmentation, and achieves the second best performance in the detection task, even though the network has been exposed only on grayscale images during pretraining. We also summarize the comparison on classification and segmentation tasks in ~\figref{fig:cls_vs_seg} which indicates that our approach learns more robust and general-purpose representations in comparison to each of the involved tasks and all the conventional methods.  


\begin{table}
\resizebox{\columnwidth}{!}{%
\begin{tabular}{l cc cc cc cc }
\hline
\bf Method                   && \bf Class. &&\bf Det.  &&\bf Segm.  \\
\hline
\hline
ImageNet~\cite{krizhevsky2012imagenet}                &&79.9     &&56.8     &&48.0 \\
Random                       &&53.3     &&43.4     &&19.8 \\
\hline
RelativePosition~\cite{Doersch15ICCV}         &&65.3     &&51.1     && -   \\
Jigsaw~\cite{Noroozi16ECCV}                  &&67.6     &&\bf{53.2}   &&37.6 \\
Ego-motion~\cite{Wang15ICCV}               &&54.2     &&43.9     && -   \\
Adversarial~\cite{Donahue17ICLR}              &&58.6     &&46.2     &&34.9 \\
Inpainting~\cite{Pathak16CVPR}              &&56.5     &&44.5     &&29.7 \\
Colorization~\cite{Zhang16ECCV}            &&65.9     &&46.9     &&35.6 \\
Split-Brain~\cite{Zhang17CVPR}             &&67.1     &&46.7     &&36.0 \\
ColorProxy~\cite{Larsson17CVPR}              &&65.9     && -       &&\underline{38.4} \\
WatchingObjectMove~\cite{Pathak17CVPR}      &&61.0     &&52.2     && -   \\
Counting~\cite{Noroozi17ICCV}                &&\underline{67.7}  &&51.4     &&36.6 \\
CDJP &&\bf{69.2}&&\underline{52.4} &&\bf{39.3} \\
\hline
\end{tabular}
}
\smallskip
\caption{{\bf Evaluation of transfer learning on PASCAL.} Classification and detection are evaluated on PASCAL VOC 2007 with testing frameworks from ~\cite{Long15CVPR} and ~\cite{girshick15ICCV15}, respectively. Both tasks are evaluated using mean average precision(mAP) as a performance measure. Segmentation is evaluated on PASCAL VOC 2012 with testing framework from ~\cite{Long15CVPR}, which reports mean intersection over union(mIU). }
\label{table:table2}
\end{table} 


\subsection{Linear Classification on ImageNet}

We test the task-generality of our learned representations on large-scale representation learning benchmarks. As proposed in ~\cite{Zhang16ECCV}, we freeze each layer of our learned features from $conv1$ to $conv5$, and initialize the subsequent unfrozen layers with random values. Then, we train linear classifiers on top of each layer on labeled ImageNet~\cite{ILSVRC15} dataset.

The result is shown in ~\tabref{table:imagenet}. ImageNet-pretrained AlexNet shows the best performance and is the upper bound in this comparison. Since our network only learns from $L$ channel, $conv1$ features suffer lack of input information, resulting in slightly lower score compared to other methods. However, it overcomes this handicap immediately from $conv2$ layer, and achieves competitive performances in higher layers. Finally, $conv4$ and $conv5$ features achieve the second best and state-of-the-art performances, respectively. 

As shown in ~\cite{Noroozi16ECCV}, the last layers of the pretrained network tend to be task-specific, while the first layers are general-purpose. In our proposed architecture(\figref{fig:architecture}), this transition from general-purpose to task-specific is delayed and left to the task branches. Since the last features of the shared tower must support all three different, they should remain as general as possible, rather than get biased to either of the tasks. Also, the network can hardly assign separate features to each tasks since the features required by the tasks often overlap, thus it has to integrate and hold the different features up to the last layers.

\begin{table}

\resizebox{\columnwidth}{!}{%
\begin{tabular}{l cc c c c c c}
\hline
\bf Method      && \bf conv1 & \bf conv2 & \bf conv3 & \bf conv4 & \bf conv5 \\
\hline
\hline
ImageNet~\cite{krizhevsky2012imagenet}               &&19.3 &36.3 &44.2 &48.3 &50.5 \\
Random                      &&11.6 &17.1 &16.9 &16.3 &14.1\\
\hline
\hline
RelativePosition~\cite{Doersch15ICCV}        &&16.2 &23.3 &30.2 &31.7 &29.6\\
Jigsaw~\cite{Noroozi16ECCV}                 &&\bf{18.2} &28.8 &34.0 &33.9 &27.1\\
Adversarial~\cite{Donahue17ICLR}             &&14.1 &20.7 &21.0 &19.8 &15.5\\
Inpainting~\cite{Pathak16CVPR}             &&17.7 &24.5 &31.0 &29.9 &28.0\\
Colorization~\cite{Zhang16ECCV}           &&12.5 &24.5 &30.4 &31.5 &30.3\\
Split-Brain~\cite{Zhang17CVPR}            &&17.7 &\underline{29.3} &\bf{35.4} &\bf{35.2} &\underline{32.8}\\
Counting~\cite{Noroozi17ICCV}               &&\underline{18.0} &\bf{30.6} &\underline{34.3} &32.5 &25.7\\
CDJP                       &&14.5 &27.2 &32.8 &\underline{34.3} &\bf{32.9}\\
\hline
\end{tabular}
}
\smallskip
\caption{{\bf Linear classification on ImageNet. } We train linear classifiers on top of each layer of the learned feature representations. We use publicly available testing code from \cite{Zhang16ECCV} and report top-1 accuracy of AlexNet on ImageNet 1000-way classification. The learned weights between $conv1$ and the displayed layer are frozen.}
\label{table:imagenet}
\end{table} 


\subsection{Comparing Combinations of Self-supervised tasks}
\label{sec:comparing}
In order to show the impact of each task, we evaluate different combinations on PASCAL classification and semantic segmentation tasks. We experiment with the same architecture~\figref{fig:architecture}, but with or without certain task branches to make different combinations. In addition, as we mentioned in ~\secref{sec:Damaged Puzzles}, we provide the result of the simple combination of the tasks in their original form, which conceptually follows ~\cite{Doersch17ICCV}.

The results are shown in ~\tabref{table:ablation}. We set the jigsaw puzzle as our starting point and add different tasks to it. We can see that the performances increase every time the tasks are combined. Our final method which combines all three tasks obtains the best scores and improves our jigsaw puzzle by 2.6\% and 2.5\%scores both in classification and semantic segmentation tasks. The simple combination of the original versions slightly improves their single-task baselines~\cite{Noroozi16ECCV,Pathak16CVPR,Zhang16ECCV} in both test tasks, but not better than our $Jig.$+$Col.$ and $Jig.$+$Inp.$+$Col.$(CDJP) methods.

\begin{table}
\centering

\resizebox{0.9\columnwidth}{!}{%
\begin{tabular}{l cc cc cc cc }
\hline
\bf Combination                   &&& \bf Class. &&&\bf Segm.  \\
\hline
\hline

Jig.                    & &&66.6    & &&36.8 \\
Jig.+Inp.              &  &&67.4    & &&37.9 \\
Jig.+Col.               & &&68.4    & &&38.6 \\
\hline
Jig.+Inp.+Col./simple    & &&68.0    & &&38.1 \\
\hline
Jig.+Inp.+Col.(CDJP)         & &&\bf{69.2}    & &&\bf{39.3} \\
\hline

\end{tabular}
}
\smallskip
\caption{{\bf Comparing different combinations of self-supervised tasks on PASCAL.} we evaluate different combinations of self-supervised tasks on PASCAL classification and segmentation in the same setting as in \tabref{table:table1}. We make different combinations using our architecture (\figref{fig:architecture}) with or without certain task branches; this may have caused slight performance differences from the original task. We also report the result of simple combinations where the original versions of each tasks are jointly trained. }
\label{table:ablation}
\end{table} 

\begin{figure*}[]
\label{fig:retrieval}
\centering
\def\arraystretch{1.0}
\begin{tabular}{@{}c@{\hskip 0.005\textwidth}c@{\hskip 0.02\textwidth}c@{\hskip 0.005\textwidth}c@{\hskip 0.001\textwidth}c}
\includegraphics[width=0.065\textwidth]{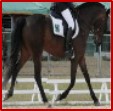} &
\includegraphics[width=0.39\textwidth]{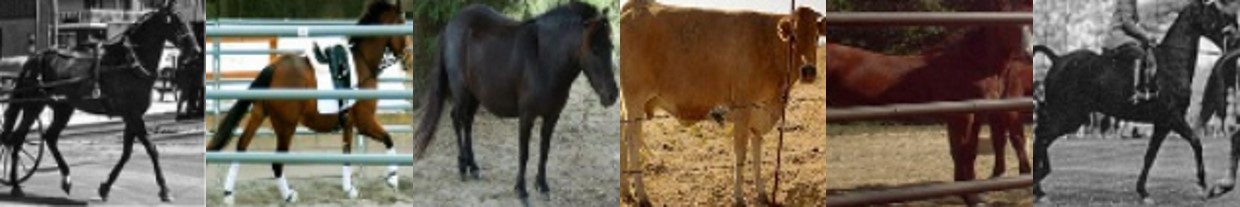} &
\includegraphics[width=0.065\textwidth]{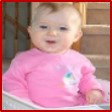} &
\includegraphics[width=0.39\textwidth]{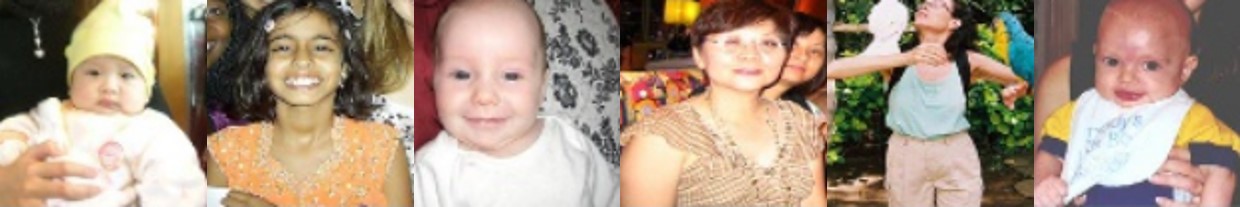} & Jig.\\ 
&
\includegraphics[width=0.39\textwidth]{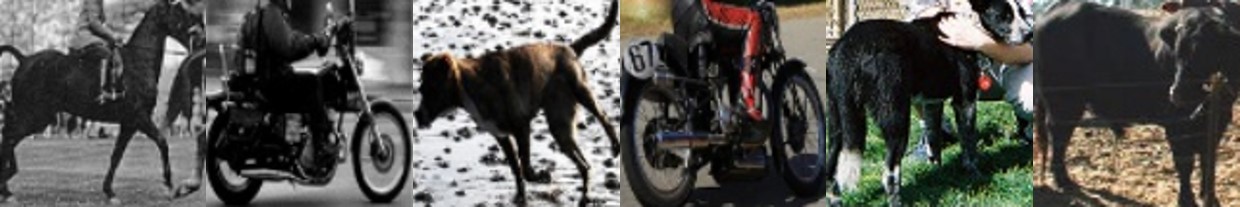} &
&
\includegraphics[width=0.39\textwidth]{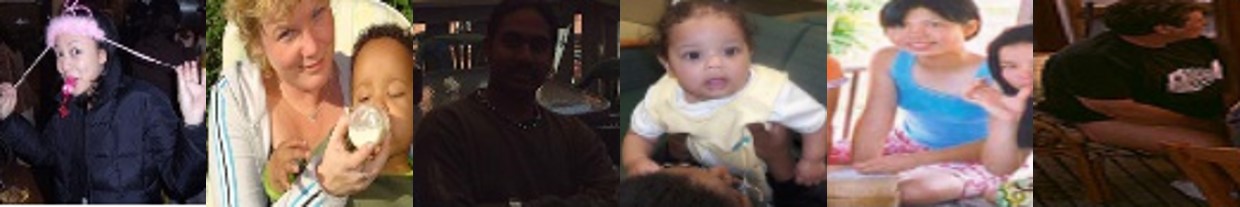} & Inp.\\
&
\includegraphics[width=0.39\textwidth]{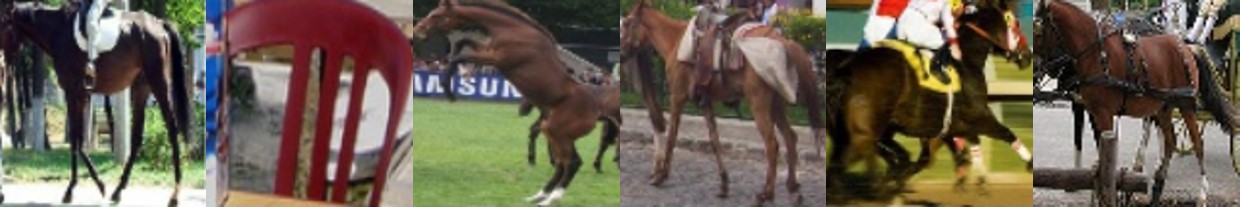} &
&
\includegraphics[width=0.39\textwidth]{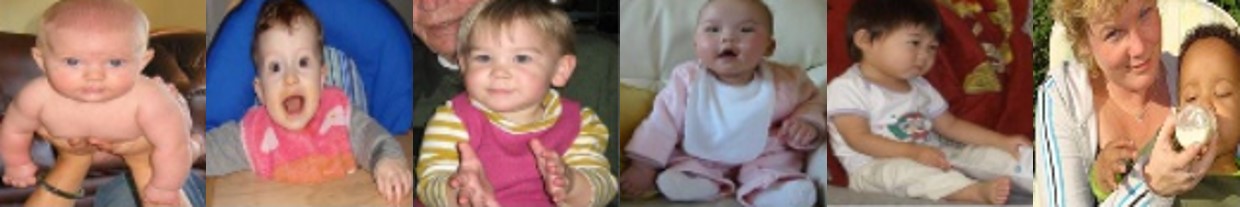} & Col.\\
&
\includegraphics[width=0.39\textwidth]{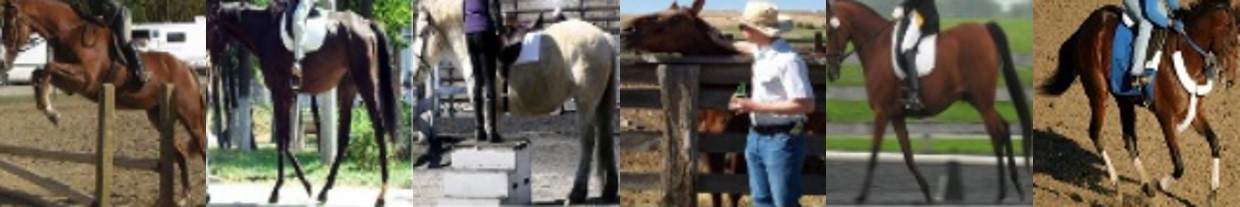} &
&
\includegraphics[width=0.39\textwidth]{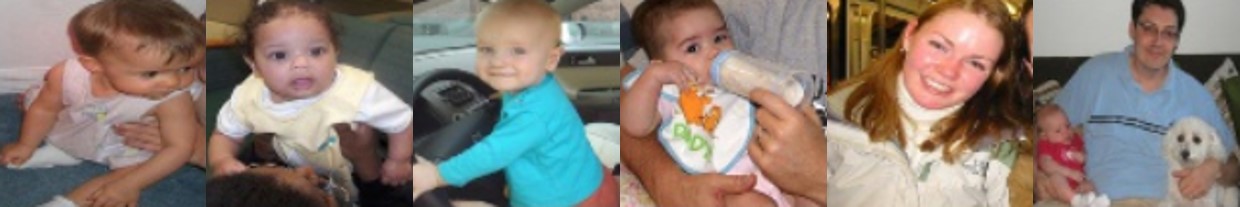} & Ours\\
&
\includegraphics[width=0.39\textwidth]{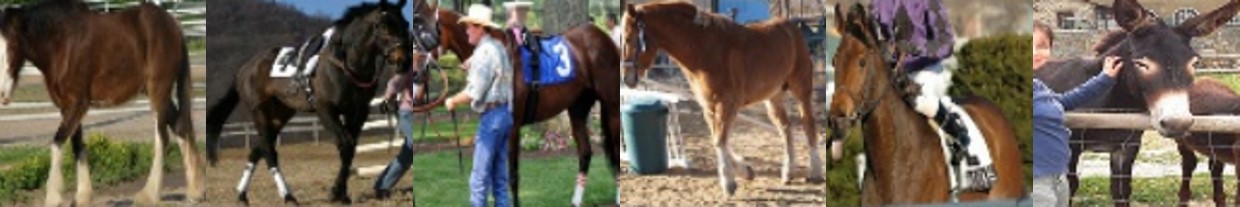} &
&
\includegraphics[width=0.39\textwidth]{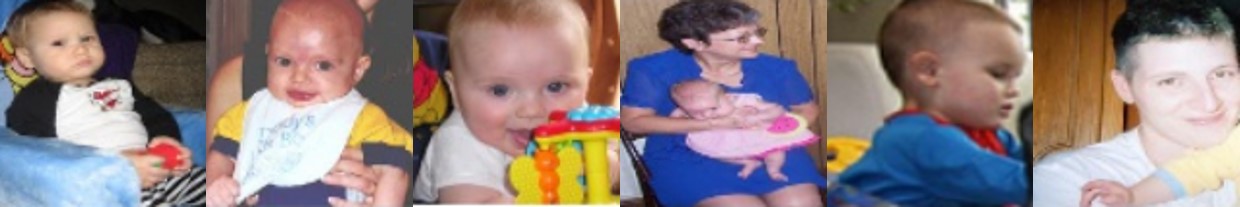} & INet\\ \\

\includegraphics[width=0.065\textwidth]{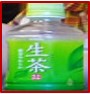} &
\includegraphics[width=0.39\textwidth]{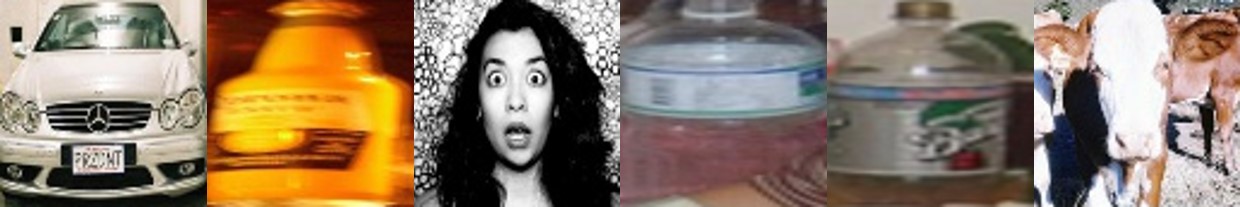} &
\includegraphics[width=0.065\textwidth]{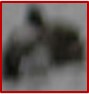} &
\includegraphics[width=0.39\textwidth]{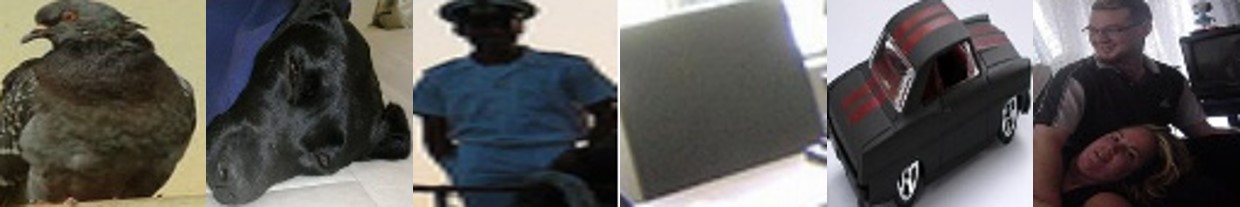} & Jig.\\
&
\includegraphics[width=0.39\textwidth]{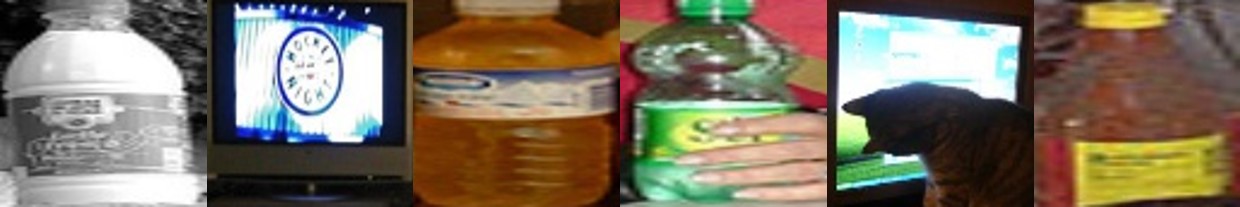} &
&
\includegraphics[width=0.39\textwidth]{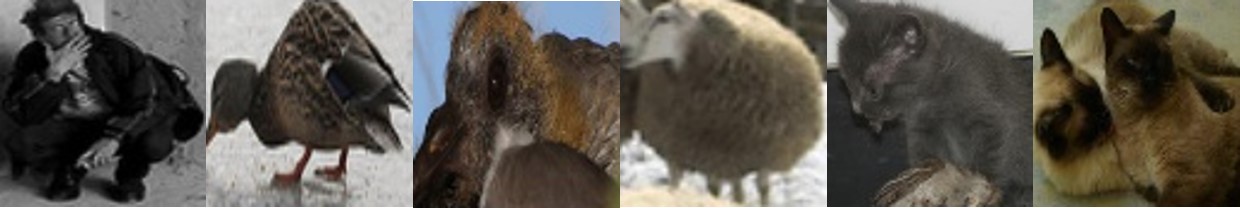} & Inp.\\
&
\includegraphics[width=0.39\textwidth]{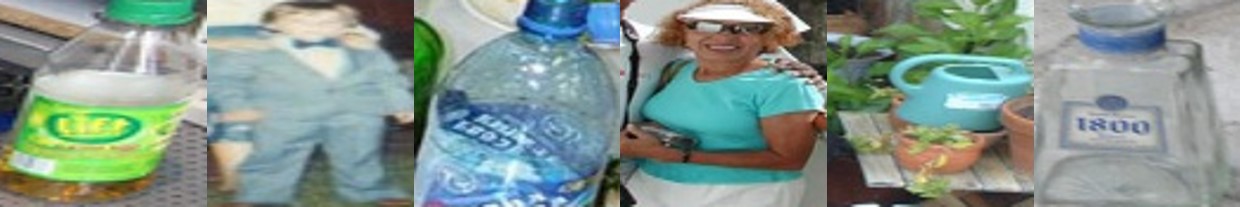} &
&
\includegraphics[width=0.39\textwidth]{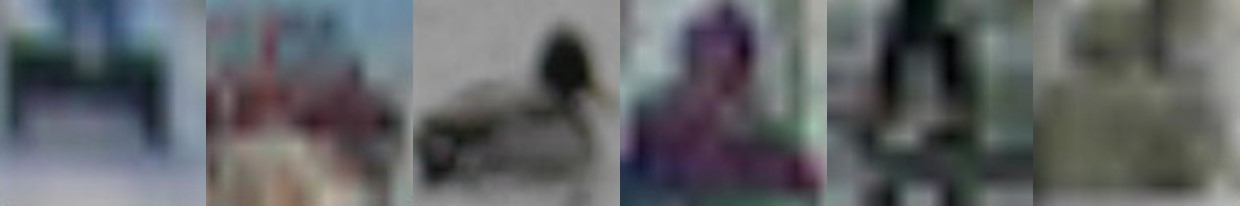} & Col.\\
&
\includegraphics[width=0.39\textwidth]{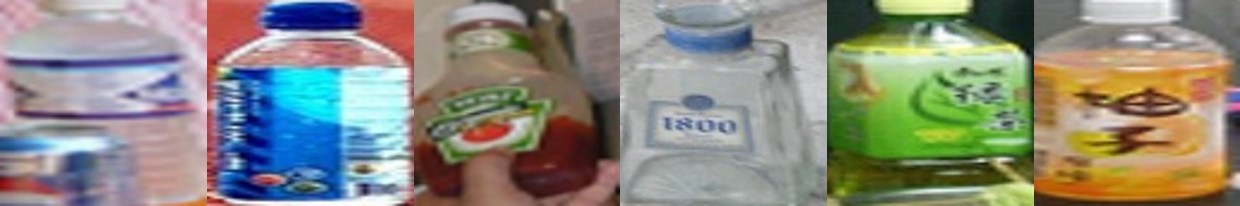} &
&
\includegraphics[width=0.39\textwidth]{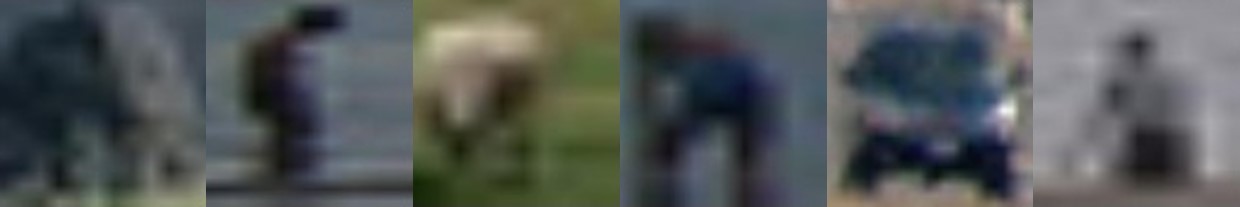} & Ours\\
&
\includegraphics[width=0.39\textwidth]{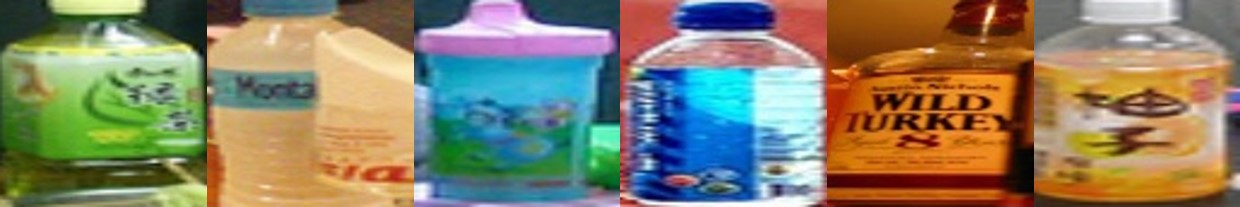} &
&
\includegraphics[width=0.39\textwidth]{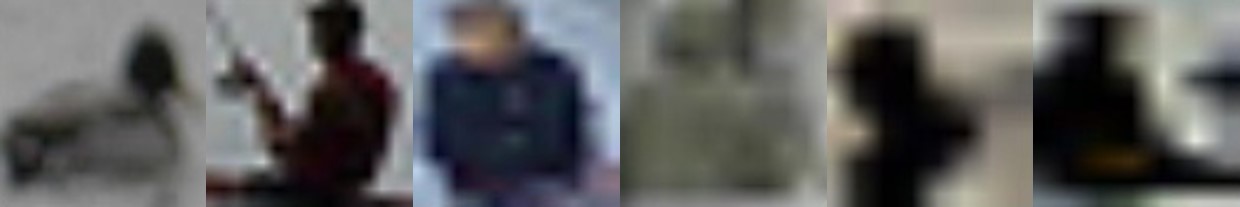} & INet\\

\end{tabular}

\caption{{\bf Nearest Neighbor Search. } We perform image retrieval on the object instances cropped from the PASCAL VOC 2012~\cite{pascal-voc-2012} \textit{trainval} dataset. The query images are in red boxes. Down from the top rows are the retrieval results of jigsaw puzzle, inpainting, colorization, our method, and ImageNet classification, respectively. }
\label{fig:nearest}
\end{figure*}

\subsection{Nearest Neighbor Search}

The pretrained networks recognize the semantic similarity of data by their own standards. We qualitatively evaluate the validity of this reasoning of the networks by performing `nearest neighbor search' which has been proposed in \cite{Doersch15ICCV} and further used in \cite{Wang17ICCV, Noroozi17ICCV}. In this experiment we compare AlexNets\cite{krizhevsky2012imagenet} pretrained by different methods: jigsaw puzzle\cite{Noroozi16ECCV}, inpainting\cite{Pathak16CVPR}, colorization\cite{Zhang16ECCV}, ours, and ImageNet classification\cite{krizhevsky2012imagenet}. We perform retrieval on $fc6$ (the feature before the concatenation) for jigsaw puzzle, $conv5$ (the last layer of the encoder) for inpainting, and $conv7$/$fc7$ features for the remaining methods.

\bigskip\noindent\textbf{Single-task Baselines. } As in figure 5, the learned representations in each methods show distinct characteristics. For example, the jigsaw puzzle representations retrieve objects with the same pose and shape. Even in the \textit{blurred} image, it retrieves objects with similar silhouettes. In inpainting, objects that would co-occur or share the similar background are retrieved, such as things to ride for \textit{horse} and caregivers for \textit{baby}. The features learned by colorization is often color-specific, and retrieves babies wearing pink clothes for \textit{baby}, and sometimes false samples with blue-green color for \textit{bottle}. Also, blurred objects a retrieved for the \textit{blurred} image. Such color-sensitivity sometimes misrepresent semantics, \textit{e.g.} a brown chair back is retrieved for \textit{horse} image.

\bigskip\noindent\textbf{Similarity to ImageNet Classification Pretraining. }
Note that we consider pretraining on ImageNet classification as our gold standard in this qualitative evaluation. Our approach integrates the characteristics of the single-task baselines, yet mostly complement and overcome the aforementioned sensitivities. First, our approach is more invariant to pose/viewpoint variations compared to jigsaw puzzle baseline, and represents \textit{horse}s and \textit{baby}s in different pose and viewpoint as semantically nearby, which is also the case in `ImageNet' model. Furthermore, our representations are more robust in intra-class color variations, and retrieves objects with various colors according to \textit{horse}, \textit{baby}, and \textit{bottle} query images, which also raise our model closer to our gold standard. Our model also adopts the virtues of the single-task baselines. To illustrate, for \textit{blurred} object, as in colorization, our model retrieves images that are semantically ambiguous. We can see the same tendency in the `ImageNet' model, where it may consider the query image to be vague, and retrieves also blurred objects in different categories. Finally, our model adopts a reasonable understanding on the image context, which enabled the retrieval of co-occurable objects, \textit{e.g.,} person with \textit{horse} and parent with \textit{baby}. Interestingly, we observe that the `ImageNet' retrieves images where person and horse; caregiver and baby appear together, similarly to ours. These results can be viewed as one reason that our approach can propagate the high-level semantics through our model, and raise its robustness and task generality of our representations.

\section{Conclusions}
In this paper, we study complicating self-supervised tasks for representation learning. We propose complicated versions of jigsaw puzzles, inpainting and colorization and show their effectiveness on representation learning. Furthermore, we design ``Completing damaged jigsaw puzzles" as a more complicated and complex problem for self-supervised representation learning. While learning to recover and colorize original image content simultaneously, rich and general-purpose visual features are encoded into the network. Experiments contain transfer learning on PASCAL VOC classification, detection and segmentation, ImageNet linear classification as well as nearest neighbor search. All of the results clearly show that the features learned by our method generalize well across different high-level visual tasks.

\paragraph{Acknowledgements}
This research is supported by the Study on representation learning for object recognition funded by the Samsung Electronics Co., Ltd (Samsung Research)
\clearpage
\clearpage

{\small
\bibliographystyle{ieee}
\bibliography{egbib}
}

\end{document}